\documentclass[letterpaper, 10 pt, conference]{ieeeconf}  % Comment this line out if 

\usepackage{graphicx}
\usepackage{threeparttable}
\usepackage{amsmath}
\usepackage{mathtools}
\usepackage{amssymb}
\usepackage{xcolor}
\usepackage{multirow}
\usepackage{caption}
\usepackage{subcaption}
\usepackage{pgfplots} % package used to implement the plot  
\pgfplotsset{width=6.5cm, compat=1.6} 

\usepackage{algorithm}
\PassOptionsToPackage{noend}{algpseudocode}% comment out if want end's to show
\usepackage{algpseudocode}% http://ctan.org/pkg/algorithmicx

\errorcontextlines\maxdimen

\makeatletter
\newcommand*{\algrule}[1][\algorithmicindent]{\makebox[#1][l]{\hspace*{.5em}\vrule height .75\baselineskip depth .25\baselineskip}}%

\newcount\ALG@printindent@tempcnta
\def\ALG@printindent{%
    \ifnum \theALG@nested>0% is there anything to print
        \ifx\ALG@text\ALG@x@notext% is this an end group without any text?
            \addvspace{-3pt}% FUDGE for cases where no text is shown, to make the rules line up
        \else
            \unskip
            \ALG@printindent@tempcnta=1
            \loop
                \algrule[\csname ALG@ind@\the\ALG@printindent@tempcnta\endcsname]%
                \advance \ALG@printindent@tempcnta 1
            \ifnum \ALG@printindent@tempcnta<\numexpr\theALG@nested+1\relax% can't do <=, so add one to RHS and use < instead
            \repeat
        \fi
    \fi
    }%
\usepackage{etoolbox}
\patchcmd{\ALG@doentity}{\noindent\hskip\ALG@tlm}{\ALG@printindent}{}{\errmessage{failed to patch}}
\makeatother

\newbox\statebox
\newcommand{\myState}[1]{%
    \setbox\statebox=\vbox{#1}%
    \edef\thealgruleheight{\dimexpr \the\ht\statebox+1pt\relax}%
    \edef\thealgruledepth{\dimexpr \the\dp\statebox+1pt\relax}%
    \ifdim\thealgruleheight<.75\baselineskip
        \def\thealgruleheight{\dimexpr .75\baselineskip+1pt\relax}%
    \fi
    \ifdim\thealgruledepth<.25\baselineskip
        \def\thealgruledepth{\dimexpr .25\baselineskip+1pt\relax}%
    \fi
    \State #1%
    \def\thealgruleheight{\dimexpr .75\baselineskip+1pt\relax}%
    \def\thealgruledepth{\dimexpr .25\baselineskip+1pt\relax}%
}

\definecolor{Yellow}{rgb}{1,0.9,0.7}
\definecolor{Pink}{rgb}{1,0.85,0.85}
\definecolor{AntiqueWhite}{rgb}{0.9,0.9,0.9}

\newcommand{\NOTE}[1]%
{
\noindent
\fboxsep=2mm\fcolorbox{black}{AntiqueWhite}{\parbox{0.95\columnwidth}
{\textbf{NOTE: } #1}
}
}

\IEEEoverridecommandlockouts                     

\title{\LARGE \bf
Energy-Efficient Mobile Robot Control via Run-time Monitoring of Environmental Complexity and Computing Workload
}

\author{Sherif A.S. Mohamed$^{1}$, % <-this % stops a space,
Mohammad-Hashem Haghbayan$^{1, 3}$,
Antonio Miele$^{2}$,
Onur Mutlu$^{3}$,
and Juha Plosila$^{1}$

\thanks{$^{1}$Sherif A.S. Mohamed, Mohammad-Hashem Haghbayan, and Juha Plosila are with Autonomous  Systems  Laboratory (ASL), University of Turku, 20500, Turku, Finland. {\tt\small \{samoha, mohhag, juplos\}@utu.fi}}%
\thanks{$^{2}$Antonio Miele is with the Dipartimento di Elettronica, Informazione e Bioingegneria, Politecnico di Milano, 20133 Milano, Italy. {\tt\small antonio.miele@polimi.it}}%
\thanks{$^{3}$Mohammad-Hashem Haghbayan and Onur Mutlu are with ETH Zürich, 8092, Zürich, Switzerland. {\tt\small omutlu@gmail.com}}%
}

\IEEEoverridecommandlockouts

\begin{document}
\maketitle
\thispagestyle{empty}
\pagestyle{empty}

%%%%%%%%%%%%%%%%%%%%%%%%%%%%%%%%%%%%%%%%%%%%%%%%%%%%%%%%%%%%%%%%%%%%%%%%%%%%%%%%
\begin{abstract}
We propose an energy-efficient controller to minimize the energy consumption of a mobile robot by dynamically manipulating the mechanical and computational actuators of the robot. The mobile robot performs real-time vision-based applications based on an event-based camera. The actuators of the controller are CPU voltage/frequency for the computation part and motor voltage for the mechanical part. We show that independently considering speed control of the robot and voltage/frequency control of the CPU does not necessarily result in an energy-efficient solution. In fact, to obtain the highest efficiency, the computation and mechanical parts should be controlled together in synergy. 
We propose a fast hill-climbing optimization algorithm to allow the controller to find the best CPU/motor configuration at run-time and whenever the mobile robot is facing a new environment during its travel. Experimental results on a robot with Brushless DC Motors, Jetson TX2 board as the computing unit, and a DAVIS-346 event-based camera show that the proposed control algorithm can save battery energy by an average of 50.5\%,  41\%, and 30\%, in low-complexity, medium-complexity, and high-complexity environments, over baselines.

\end{abstract}

%%%%%%%%%%%%%%%%%%%%%%%%%%%%%%%%%%%%%%%%%%%%%%%%%%%%%%%%%%%%%%%%%%%%%%%%%%%%%%%%

%https://arxiv.org/pdf/1811.10277.pdf
\section{Introduction}

%\NOTE{this intro is too long and too vague. It should go to the point faster and discuss the main issue that is the limited power/energy budget. Moreover, the intro lacks of references.... }

Constrained energy is a major challenge in the field of mobile robotics. The research focus has been mainly on optimizing the \textit{motion planning} of a robot and \textit{kinematic energy}~\cite{b2,AAA}. However, kinematic energy consumption is not the only source of energy drain. A mobile robot, as a cyber-physical device, also contains a \textit{cyber-part} beside the \textit{physical part}, such as on-board electrical devices, micro-controllers, and sensors, each of which consumes power and contributes to the overall energy consumption~\cite{battery-1}. For example, the execution of heavy vision-based applications that use onboard computing units and sensors drains a significant portion of the available source of energy. \textit{Computing energy consumption} of the cyber-part in a robot is one of the significant portion of the energy consumption in a robot that should be considered while optimizing the energy.

There have been several studies to reduce the power/energy consumption of the computing units for cyber-physical systems (e.g.,~\cite{c4,c1,c2,c3}). In recent mobile robots computing power management units significantly reduces the computing power/energy consumption via dynamic voltage and frequency scaling (DVFS), power gating (PG) and/or other hardware knobs, whenever full performance is not required (e.g.,~\cite{DVFS, iccd-2014,antonio,d1,d2}).

Prior techniques try to improve the power/energy consumption of the cyber-part and/or physical-part of the robotic system separately and independent from each other, without considering the co-relations between these two parts. One of these co-relations is the effect of robot mechanical \textit{control decision} on the \textit{computing workload}. In a mobile robot with a normal frame-based camera as a sensor, the quality of an image of a scene captured at different speeds is not the same~\cite{ektl,forster2015continuous}. This results in \textit{workload variation} for processing different captured images of \textit{the same scene} at \textit{different speeds} of the robot\footnote{This happens due to changes in the captured image quality.}. Another example is a mobile robot with an \textit{event-based camera} as the sensor. Here, both the \textit{environmental complexity} and \textit{the robot speed} affect the number of generated events at each instance of time and consequently the event processing workload~\cite{entropy, sc2}. 

%together affects the workload~\cite{entropy}. Since the speed of the robot is determined by the mechanical controller of the mobile robot, it can be concluded that, while using e.g. event-based camera in a mobile robot, in addition to the complexity of the environment, the decision in robot's mechanical controller also affects the workload of the computational units. 

\begin{figure}[t]
         \centering
         \includegraphics[width=.47\textwidth]{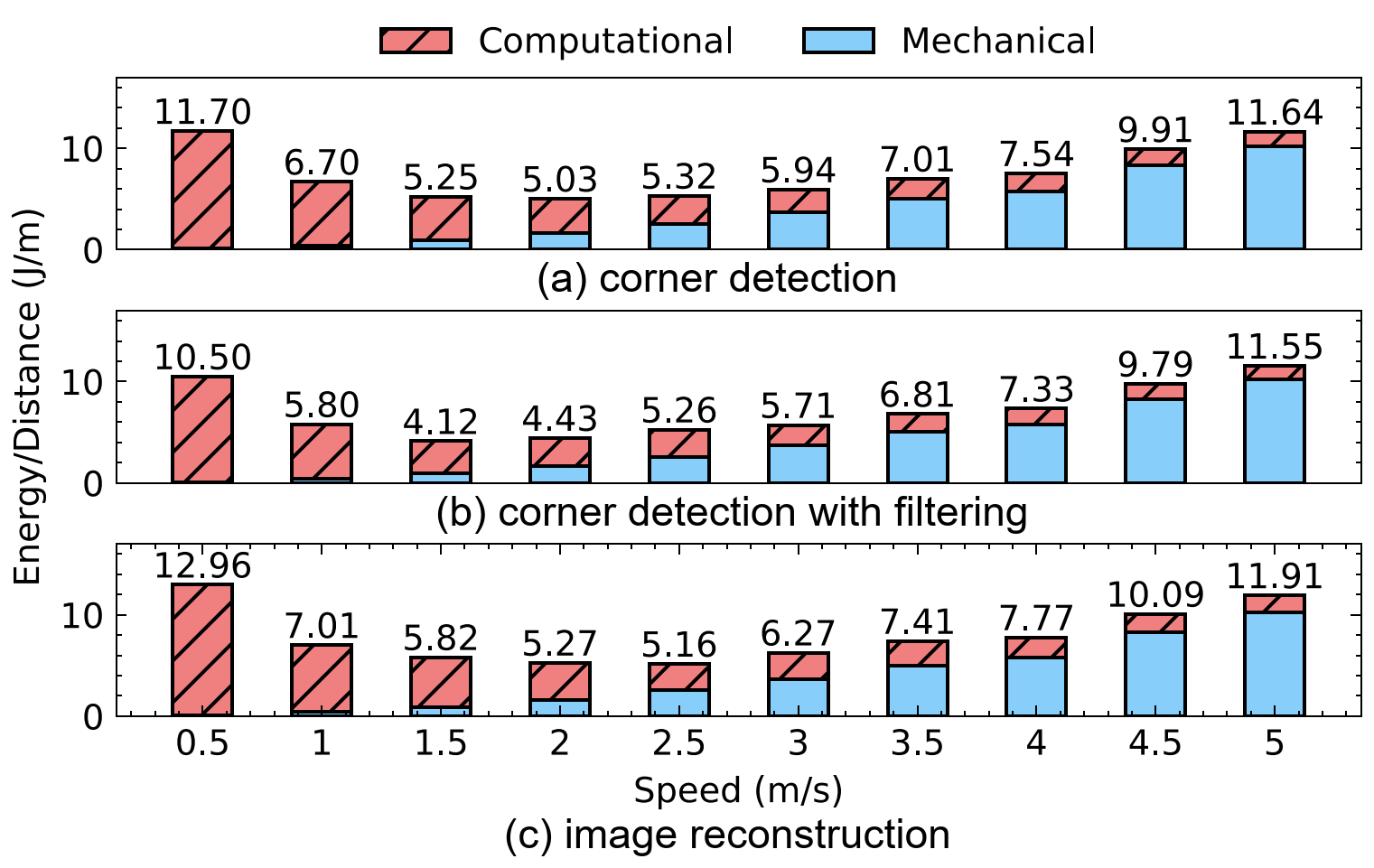}
         \caption{Comparing average power consumption of computational and mechanical parts of a mobile robot for three vision-based applications at various robot movement speed. %\textcolor{blue}{: corner detection, corner detection with filtering, and intensity-based image reconstruction, respectively.}
         }
        \label{fig:cpu_motor_power}
        \vspace{-10pt}
\end{figure}

As an example, Figure~\ref{fig:cpu_motor_power} shows the energy consumption per unit distance (referred to as \textit{locomotion cost} in literature \cite{EPD, LM1, LM2}) for different speeds of a mobile robot with an event-camera sensor, when running three different image processing tasks on the event data\footnote{Details on our applications are provided in Section~\ref{sec:appl}}. The Power Management Unit is responsible for reducing the computing power consumption as much as possible, while avoiding unacceptable throughput loss, by manipulating the voltage and frequency of the processing unit. 
%In particular, in each plot a different application\footnote{Details on the applications are provided in Section~\ref{sec:appl}} capturing visual information through an event-based camera and performing vision-based functionalities is executed, and the same activity is repeated on the same path at different speeds.}
%Figure~\ref{fig:cpu_motor_power} shows the energy consumption per unit distance in a mobile robot running one of three different feature detection methods as vision-based applications\footnote{Details on the applications are provided in Section~\ref{sec:appl}}. The overall goal of the robot is to capture the visual information through an event camera and perform vision-based application, e.g., corner detection. The source and destination of the robot for all the experiments are the same. 
The overall drained energy of the robot is calculated as follows:
\begin{equation}
E = \int^T_0 \left(p^v_{motor}+p^{e}_{cpu}\right) dt
\end{equation}
where $p_{motor}$, $p_{cpu}$ are respectively the power of the mechanical motors, $motor$, and computing unit, $cpu$, which are functions of robot speed, $v$, and captured events, $e$, respectively. The integral spans over the entire duration of the mission $T$ to compute the energy consumption.

%, $v$, $e$, and $t$ are the average power of the motors, the average power of the board, robot's speed, the complexity of the environment, and duration of the mission, respectively. 
Figure~\ref{fig:cpu_motor_power} shows that increasing the speed to the maximum possible value does not necessarily result in an optimal energy consumption. In fact, if the robot's mechanical controller uses the \textit{highest possible speed} to reduce the movement time, in addition to the requirement for high instantaneous power of the motors, computing a \textit{large number of generated events} per time unit consumes a high amount of power. Having a low speed for the robot saves mechanical energy significantly. On the other hand, computing energy becomes dominant despite a lower frequency of generated events. The reason for this is that, by decreasing the speed and increasing the overall time of the travel, most of the energy consumed by the processing units is \textit{static energy} that is used constantly during the waiting periods of the units. It can be seen that the lowest energy consumption is obtained at \textit{a specific speed} of the robot. Another important fact that can be observed is that the most energy-efficient speed for different applications are not necessarily the same value. This is due to the different strategies in the Power Management Unit for different types of software executing on the computing platform. Therefore, to obtain energy-efficient control of the robot, both computational and mechanical controllers must be tuned \textit{together at run-time}, to dynamically adjust the robot speed and voltage/frequency of the computing unit. %In particular, in this example slowing down the speed of the motors and tuning the computational power of the CPU, by means of DVFS, to be able to process the stream of incoming events allows to obtain an energy saving approximately of the $50\%$.
%This is because of this fact that an increase in the speed of the robot not only exponentially increases the power consumption of the motors but also increases the power consumption of the processing units because of the computation of an enormous number of generated events in each instance of time. Putting these all together, it can be concluded that, 
%Another fact that can be observed in Figure~\ref{fig:cpu_motor_power} is that the best energy-efficient solution for each application might not be the same as the other. 

Given the above motivation, this paper presents a novel energy-efficient control technique to manipulate the mechanical and the computational actuators of a mobile robot together to obtain the best possible overall energy consumption. The actuators are the voltage of the motors, to change the speed, and the voltage/frequency of the CPU. To manipulate the actuators, the instantaneous power of the mechanical and computational parts are fed back to the controller. Then, the controller performs a fast hill-climbing algorithm at run-time to \textit{iteratively improve} the system configuration to minimize the overall energy consumption.

In summary the novel contributions of the paper are:

\begin{itemize}
\item A controller framework capable of co-managing mechanical and computational parts of the robot system executing vision-based applications.
\item A control policy to select the best motor speed and CPU voltage level configurations to guarantee application throughput while minimizing energy consumption.
\item An experimental session showing an improvement in energy saving up to 50\% w.r.t. the baseline solutions.
\end{itemize}

The rest of the paper is organized as follows. Section~\ref{sec:background} provides the working scenario and the motivation for the proposed controller. In Section~\ref{sec:approach}, the proposed approach is described. Section~\ref{sec:results} provides the experimental results. Section~\ref{sec:relatedworks} is devoted to the related works. Finally, Section~\ref{sec:conclusions} concludes the paper with some discussion and comments on future work.

\section{Working scenario and motivation}\label{sec:background}
We present the overall robot architecture and a motivating example that shows how in different environmental situations the most energy-efficient values for voltage/frequency and robot speed changes. This motivational example demonstrates an overview of the dynamics of the system in different \textit{environmental complexities} and motivates the proposed approach for an energy-efficient run-time controller.

\subsection{Working scenario}

\noindent\textbf{System architecture.}
The platform in this work consists of three main parts, i.e., overall robot architecture, the architecture of the computing unit, and vision-based applications. Each of them is separately explained below.  

\begin{figure}[t]
    \centering
    \includegraphics[width=.32\textwidth]{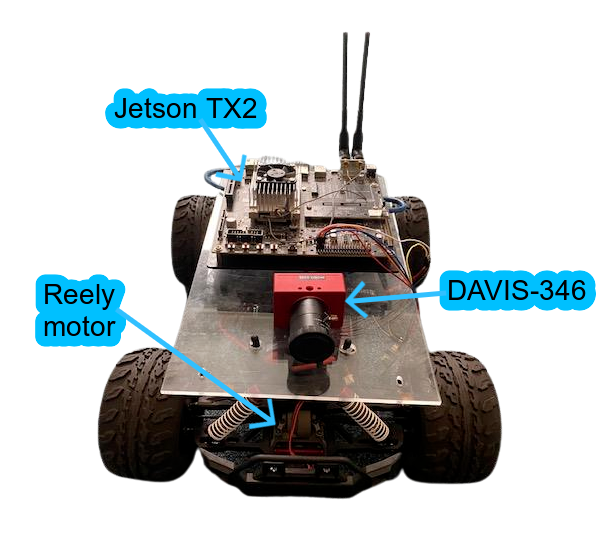}
    \vspace{-7pt}
    \caption{Platform setup.}
    \vspace{-10pt}
    \label{fig:ps}
\end{figure}

\begin{figure*} [tb]
\vspace*{.2cm}
     \centering
     \begin{subfigure}[b]{.32\textwidth}
         \centering         \includegraphics[width=1\textwidth]{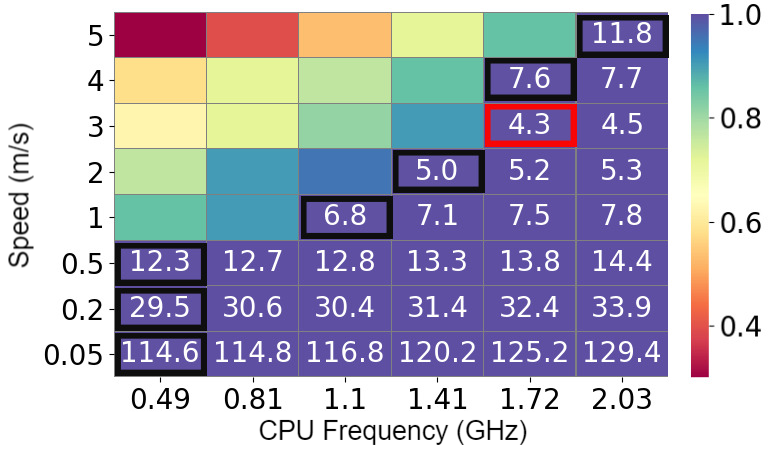}
         \caption{Low-complexity environment}
         \label{fig:1}
     \end{subfigure}
     \begin{subfigure}[b]{0.32\textwidth}
         \centering
         \includegraphics[width=1\textwidth]{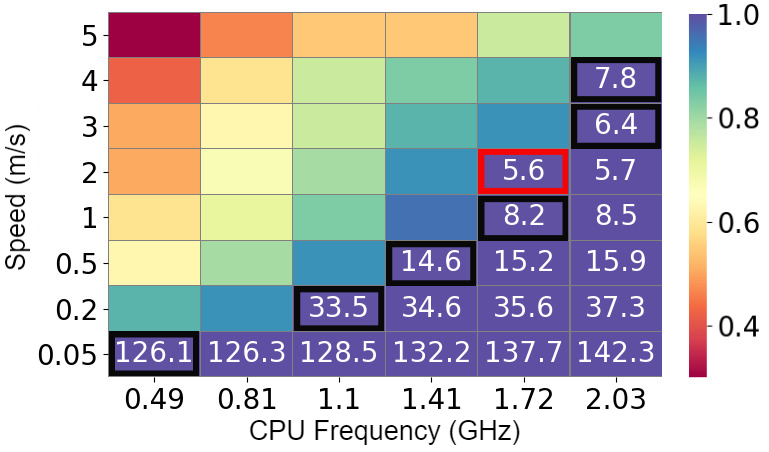}
         \caption{Medium-complexity environment}
         \label{fig:2}
     \end{subfigure}
     \begin{subfigure}[b]{0.32\textwidth}
         \centering
         \includegraphics[width=\textwidth]{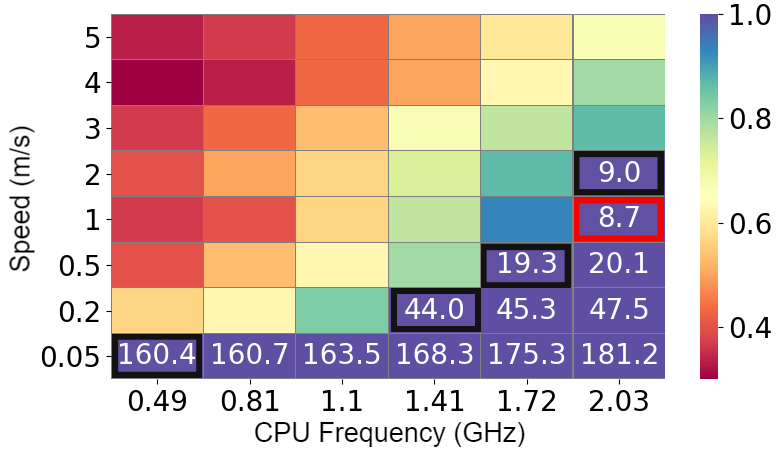}
         \caption{High-complexity environment}
         \label{fig:6}
     \end{subfigure}
        \caption{The parameters configuration space for the robot while running three applications with different environmental complexity. The color of each cell represents the achieved application throughput (spanning from 0 to 1), while for valid configurations (i.e. with throughput equal to 1) the reported number is the energy consumption in Joule.}
        \label{fig:res}
\end{figure*}

\noindent\textbf{Overall robot architecture.}
This includes the physical parts of the platform, such as an embedded system, an event-based camera, and a brushless dc motor, as shown in Figure~\ref{fig:ps}. The Jetson TX2 board is a super power-efficient embedded board with two clusters: quad-core ARM Cortex-A57 and dual-core NVIDIA Denver. The event-based camera, i.e., DAVIS-346, can capture both intensity images and a stream of asynchronous events with a high temporal resolution of up to 10 million events in a second~\cite{eventSurvey, davis}. %Event cameras require fundamentally different methods than traditional cameras to process their data, since they publish events instead of intensity images. Events are encoded as 65-bits. The 2D position (x,y) of the pixel in the sensor is presented. The 32-bit timestamp identifies when the event occurred. A 1-bit flag is used to represent the polarity (0,1) of the event. The polarity indicates the change of the brightness. 
A DC brushless motor is used to power up the robot. The motor has 3000 kV which can spin at 50000 RPM. The power supply of motors is calculated based on the current measurement circuit that measures the instantaneous electric current toward the motors.

\noindent\textbf{The architecture of the CPU controller.}
For executing the applications, a mapping unit, as the middleware proposed in~\cite{antonio}, allocates the tasks on the various cores inside the CPU clusters. The cores can run at a maximum frequency of 2 GHz. However, run-time DVFS is capable to reduce the voltage/frequency of the CPU cluster at various intermediate steps down to 300 kHz. The middleware is capable to report the throughput for each application that helps to show if the application is loosing some parts of the data due to overload or not. Throughput value ranges between 0 to 1 where reported value 1 means all the captured data are fully being processed.

%Recent studies have shown that there should be a thorough management of different requirements and constraints across the operating system, compiler, and architectural layers~\cite{tc-2017, tvlsi-2017-2}. To have a better assessment about the current state of the resources, in terms of current power consumption, temperature, and so on, several monitoring features are mounted on the resources helping the management unit by providing frequent information about specific resources in run-time. \NOTE{How do we calculate the energy} The computational capability and power of the processing units can be varied in the modern computing platform. Dynamic voltage frequency scaling (DVFS) and per-core power gating (PCPG) are two well-known features that can restrict and decrease the computation capability and power of the processing units in run-time. In mobile robots, that perform vision-based applications, if the battery power and the capability of the computing units are not constrained, the computational operations can operate in full speed/power that results in processing the vision data for a highest possible amount. However, since the power of computation significantly contributes to the overall power consumption drained from the battery; in order to optimize the overall energy consumption of the robot, a computational controller must be aware of the computing units, to adjust the computation capability, and corresponding power, based on the required throughput and the current amount of information to process.

\noindent\textbf{Application.}\label{sec:appl}
Three different applications with different levels of complexity are used, that are image reconstruction, corner detection, and corner detection with filtering. The image reconstruction application~\cite{img_rec} is considered to have low complexity and it can produce intensity-based images from a pure event stream using a high-pass filter. The second application is a corner detector~\cite{s_ISVC}, which extracts corners from a stream of events. The application is considered to have a high level of complexity since it requires computing the eigenvalues for each incoming event to decide whether the event is a corner or not. The corner detection application is used with a three-level filtering~\cite{s_ICPR} to reduce the computational complexity and obtain a medium level of complexity.   

\begin{figure}[t]
         \centering
         \includegraphics[width=.4\textwidth]{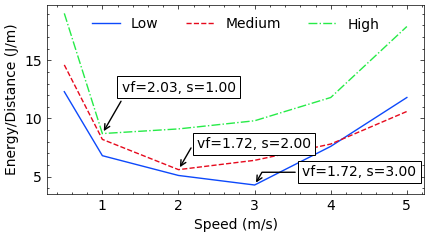}
         \caption{Energy consumption per meter of the robot in three different environmental complexities, i.e, low, medium, and high-complexity.}
         \label{fig:et}
             \vspace{-10pt}

\end{figure}

\subsection{Motivating discussion}\label{sec:motivation}
Figure~\ref{fig:res}(a-c) shows different values of the robot's overall energy and throughput levels while applying different (CPU voltage/frequency, robot speed) pairs for three kinds of environments, i.e., 1) low-complexity, 2) medium-complexity, and 3) high-complexity. It is worth noting that the throughput and energy of the mobile robot are measured for each point at run-time by enforcing the specific configuration.
The vertical axis in the figure shows the speed of the robot and the horizontal axis shows the CPU voltage/frequency level\footnote{Figures report only the frequency values since it is the knob actually tuned by the operating system; then the hardware DVFS controller selects the corresponding voltage level.}. The color of each (CPU voltage/frequency, speed) point shows the throughput of the applications, where the value 1.0, i.e., dark purple color, represents full throughput. 

As it can be noticed from the colors in the figure the 3D surface representing the throughput has a hill shape where the only global peak is placed on the right-bottom side of each plot: in fact, higher CPU frequencies provide larger computation power and lower motor speeds reduce the number of events to process per unit of time. Moreover, throughput saturates to 1.0 since the application is designed to do not over-perform when enforcing higher CPU frequency levels or lower motor speed. Therefore, the top of the hill presents a plateau of purple points.
When analyzing the three graphs together it can also be noticed that plateaus have not the same extension but at the opposite with the increase of the environmental complexity, the plateau shifts to the right-bottom side.
This comes from the fact that the number of the generated events increases and more powerful processing is needed for obtaining real-time full-throughput outcomes. 

We show inside each box the overall energy consumption of the robot. Since only the full-throughput cases are acceptable, energy values are reported for the areas with full-throughput, i.e., dark purple points.
Here there is an opposite trend w.r.t. the throughput one. In fact, energy consumption decreases when motor speed is higher (since the overall trip takes a shorter time) and the CPU frequency is lower since consuming less static power consumption. Therefore, the energy surface forms a slope in the left-top direction.

Given these considerations, we state that the most energy-efficient (CPU voltage/frequency, robot speed) pair should be found at the frontier of the purple area.
Figure~\ref{fig:et} reports only the points on the frontier of the plateau in each one of the three cases. In particular, each line is composed of the points highlighted with the black box in Figure~\ref{fig:res} (without loss of generality, to simplify the chart, we took a single point for each speed value).
From the plot, we confirm the trend, and in particular the fact that there is a single global minimum in each case and no other local minima.
Moreover, since the plateau and the corresponding frontier change with the environmental complexity, it is not possible to determine at design time a unique best configuration for any working scenario/application.
As a consequence, we here propose a control approach capable of optimizing the robot configuration at run-time by properly exploring the peculiarities of the considered problem space.

\section{The proposed controller}\label{sec:approach}

\subsection{Overall organization of the controller}
\begin{figure}[tb]
\vspace*{.2cm}
    \centering
    \includegraphics[width=0.85\columnwidth]{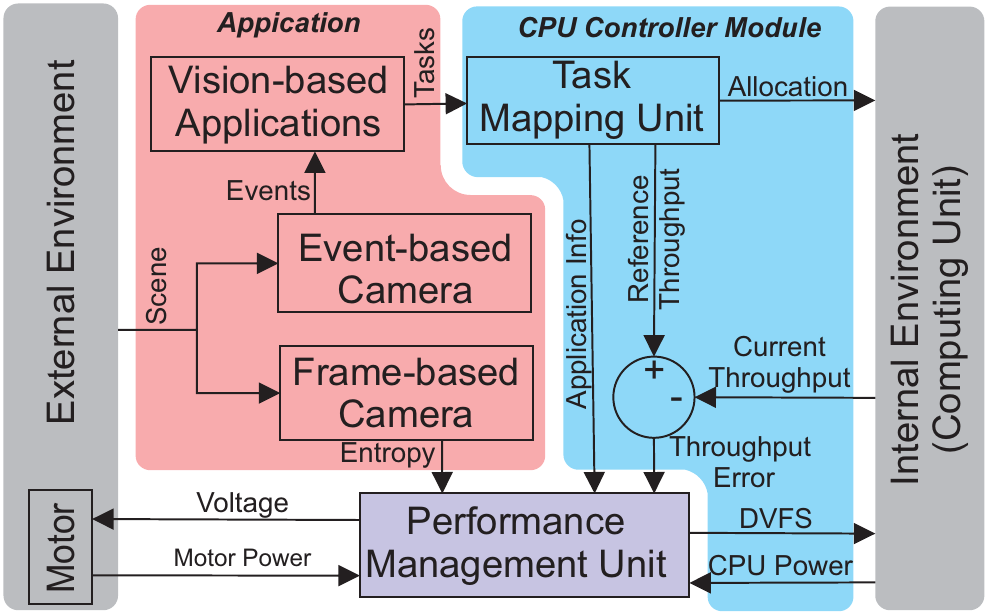}
    \caption{The system architecture of the proposed cyber-physical automation for robots.}
    \label{fig:sys}
        \vspace{-7pt}

\end{figure}

The overall structure of the proposed throughput-aware run-time controller is depicted in Figure~\ref{fig:sys}. The unit is hosted on the Performance Management Unit which controls the system based on the feedback from the Internal Environment, which is the computing unit, and External Environment, which is the environment of the mobile robot. The entropy of the vision data has been used as the metric for environmental complexity. The entropy is computed from intensity images captured by a frame-based camera. 

We borrowed a fairly standard internal organization (as in~\cite{antonio}) for the application mapping unit. Different types of applications, e.g., image reconstruction and corner detection, are given to the Application Mapping Unit in run-time for execution. Each application has a pre-specified throughput requirement that is given to the mapping unit at design time. The Application Mapping Unit allocates the applications on the CPU cluster (ARM Cortex-A57 or NVIDIA Denver) selected as the best one at design time (future work is devoted to the automatic selection of the CPU cluster at run-time without any pre-profiling). This unit also provides information about the current running applications, \textit{Application info}, and the required throughput for each application, i.e., \textit{Throughput error} $\lambda_{e}$, for the Performance Management Unit. The observations for the Performance Management Unit are 1) current power consumption of the CPU, 2) current power of the robot motors, 3) entropy of the vision sensors, and 4) the difference between the current throughput of the applications and the desired throughput, i.e., \textit{Throughput error}. Based on the feedback received from the external and the internal environments, the Performance Management Unit regulates the actuation values of the CPU voltage/frequency and motor voltage.  

During the system operation, run-time events are produced via the Event-based Camera and are passed to the Computing Unit. Based on the types of running applications, the Computing Unit processes the events. The number of events relates to two factors: 1) the environmental complexity and 2) the rate of the change in the environmental information. The environmental complexity is measured through the entropy of the captured frame-based image in run-time. The rate of the change in the environmental information is measured by the speed of the robot, which is calculated by measuring the acceleration force toward the motors. If the number of generated events in the camera increases, that might be due to the increase in the complexity of the environment or the speed of the robot, the CPU load increases that affect the throughput. The current throughput of the application must always be higher than a pre-specified throughput for the application. In the case that throughput goes below the threshold due to a heavy workload, the Performance Management Unit is capable to increase the speed of the computation by increasing the voltage/frequency of the computational cores, or, decreasing the speed of the robot by decreasing the motor voltage.

\subsection{The control algorithm}
The Performance Management Unit works as a closed-loop controller to adjust the performance based on the feedback it gets from the power sensors on the motor and the CPU. 
Due to the peculiar characteristics of the problem space discussed in Section~\ref{sec:motivation}, we have designed the controller algorithm by means of the hill-climbing algorithm, a simple and fast heuristic to search in a solution space based on local moves. Algorithm~\ref{alg:hill_climbing} shows the controller algorithm. The inputs of the algorithm are the entropy of the environmental information $et$, current power consumption of the CPU $p_{cpu}$, current power consumption of the motor $p_{motor}$, throughput error $\lambda_e$ and the application info $ai$. The outputs of the algorithm are the CPU voltage/frequency, $vf$, and the motor voltage $mv$.%, which are the actuations toward the internal and the external environments of the robot respectively, i.e., CPU and motor. %The inputs/outputs of the algorithm are compatible with the input/output signals, as shown in Figure~\ref{fig:sys}. 

The algorithm is executed whenever working scenario changes, due to entering of new applications or leaving of currently executing ones. At the beginning, the CPU voltage/frequency and motor voltage are determined based on an initialization process (Lines 1-3). In this phase, the CPU voltage/frequency and the motor voltage are adjusted based on some \textit{safe values} that approximately satisfies the required throughput; a conservative choice for our experiment is the maximum voltage/frequency level. The current entropy, $et_{current}$, is initialized to zero to be updated later in the algorithm. This initialization of entropy to zero is due to the fact that in zero entropy no event will be generated by the event-based camera and the best possible adjustment of actuations can be calculated in the initialization phase. Therefore there is no need for any optimization. However, if the captured entropy value from the environment is above zero, the process of tuning the CPU voltage/frequency and the motor voltage starts and continues until the application information is valid (Lines 4-19). During this period, whenever the input entropy, $et$, differs from the current entropy, $et_{current}$ (i.e., IF statement at Lines 5-18), the algorithm tries to find the optimal state of the system, i.e., CPU voltage/frequency and motor voltage, to minimize the overall power consumption. This ensures optimized actuations whenever the environmental complexity changes. If so, the current entropy, $et_{current}$, gets updated (Line 6), and the hill-climbing process in the 2D domain of CPU voltage/frequency and the motor voltage starts to find the optimal state (Lines 7-16). First, the neighbor states of the current state are determined (Line 8). After that, the algorithm systematically \textit{tests} all the possible neighbors of the current (CPU voltage/frequency,  motor voltage) pair to find the energy-efficient values. For each test, the new state will be applied to the robot, both CPU voltage/frequency and motor voltage, by \textit{Apply} routine (Line 10). 

\begin{algorithm}[t]
\footnotesize
  \caption{Performance Management algorithm}\label{alg:hill_climbing}
  \textbf{Input:} Entropy: $et$; CPU power: $p_{cpu}$; Motor power: $p_{motor}$; Throughput error: $\lambda_{e}$; Application Info: $ai$;\\
\textbf{Output:} CPU voltage/frequency: $vf$; Motor voltage: $mv$;\\
\textbf{Constant:} Entropy threshold: $et_{th}$; Energy threshold: $e_{th}$;\\
  \begin{algorithmic}[1]
  \State {$(vf, mv) \gets$ Initialization($ai$)}
  \State {$state_{current} \gets (vf, mv)$}
  \State {$et_{current} \gets 0$}
      \While{$ai$ is valid}
        \If{$|et_{current}-et| > et_{th}$}
            \State {$et \gets et_{current}$}
            \Repeat
                \State {$neighbors_{state}^{list} \gets$ Get\_neighbors ($state_{current}$)} 
                \For{all $(vf, mv)$ in $neighbors_{state}^{list}$} %[\{moveState\}]
                    \State {Apply ($vf, mv$)}
                    \State {Delay()}
                    \State {$E_{new}$ $\gets$ Compute\_energy($p_{cpu}$, $p_{motor}$)}
                    \If{$E_{new} - E < e_{th}$ and $\lambda_{e}$ = 0}
                        \State {$state_{new}$ $\gets$ $state$}
                        \State {$E \gets E_{new}$} 
                    \EndIf
                \EndFor
            \Until{$state_{current} = (vf, mv)$}
            \State {$state_{current} \gets (vf, mv)$}
        \EndIf
        \State{$ (vf, mv) \gets state_{current}$}
        \State Apply ($vf$,$mv$)
      \EndWhile
  \end{algorithmic}
\end{algorithm}

Then, the algorithm waits for a pre-specified \textit{delay} to be able to measure stable values for the instantaneous power of the CPU and motor and the throughput of the application. In \textit{$compute\_energy$} routine, the new overall energy consumption is computed via instantaneous power feed-back from the Computing Unit and Motor, i.e., $p_{cpu}$ and $p_{motor}$. Subsequently, the new computed overall energy consumption, $E_{new}$, is compared with the old measured energy value, $E$, to check whether the new test pair results in a less overall energy consumption or not. During this phase, throughput also should be checked since the climbing process might have a negative effect on throughput (Lines 13-15). Finally, the state that acquires the minimum obtained overall power will be considered as the new state. This process continues until the newly obtained state, $state_{new}$, is the same as the current state, $state_{current}$, which indicates that the hill-climbing process cannot improve the power consumption any more (Line 16).

It is worth noting that, even if simple, the proposed approach is effective, as also demonstrated later by the experimental results, due to the peculiar characteristics of the problem space discussed in Section~\ref{sec:motivation}. Moreover, no scalability issues may raise as the problem space cannot grow sensibly for different system architectures. In fact, the number of motor speed steps and voltage/frequency levels in a general scenario is in the order of the considered working scenario. Nonetheless, we claim that the test of all the local moves, i.e., Line 9-15, may be replaced with a smarter selection of the neighbors of the current configuration based on the slope trend shown in Figure~\ref{fig:res}. This improvement is left as a future work.

\begin{figure}[t]
\centering
\includegraphics[width=.4\textwidth]{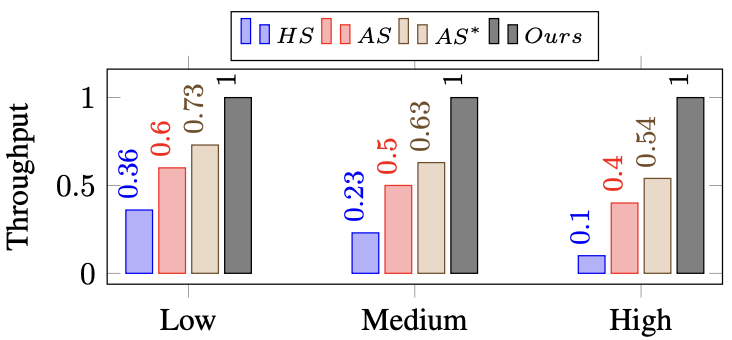}
% \resizebox{.85\columnwidth}{!}{
% \begin{tikzpicture}
% \begin{axis}[ x tick label style={
% 	/pgf/number format/1000 sep=}, height=4cm, width=8.5cm, ybar, enlargelimits=0.18, %every node near coord/.append style={font=\normalsize}, 
% 	legend style={at={(0.5,0.1)}, anchor=north,legend columns=-1, font=\fontsize{7}\selectfont}, symbolic x coords={Low, Medium, High}, ylabel= Throughput, ylabel near ticks, xtick=data, bar width=17pt, nodes near coords, every node near coord/.append style={rotate=90,anchor=south west,
%     inner ysep=0.5pt, font=\small}, 
% 	%nodes near coords align={vertical},
% 	bar width=0.3cm, legend style={at={(0.5,1.27)}}]
% % \addplot coordinates {(With TDP,0.015) (With TSP,0.02)};
% % \addplot coordinates {(With TDP,0.02) (With TSP,0.02)};
% % \addplot coordinates {(With TDP,0.007) (With TSP,0.009)};
% % \addplot coordinates {(With TDP,0.003) (With TSP,0.0037)};
% % \legend{PGCapping, MOC, RA-MOC without RB, RA-MOC}
% \addplot coordinates {(Low, 0.36) (Medium, 0.23) (High, 0.1)};
% \addplot coordinates {(Low, 0.6) (Medium, 0.5) (High, 0.4)};
% \addplot coordinates {(Low, 0.73) (Medium, 0.63) (High, 0.54)};  
% \addplot coordinates {(Low, 1) (Medium, 1) (High, 1)};  
% \legend{$HS$, $AS$, $AS^*$, $Ours$} 
% \end{axis}
% \end{tikzpicture}}
\caption{Different obtained throughput for different control techniques in three different environmental complexities: low-, medium-, and high-complexity.\label{fig:thru}}
\vspace{-15pt}
\end{figure}

%It is worth noting that a hill climbing algorithm is used to find the optimal solution since the solution space is small, i.e, the number of VF steps and the motor speed levels are limited. A fast and simple hill-climbing algorithm is suitable for small solution spaces.

%The optimization process of performance management algorithm is explained in Figure ... through an example. The figure shows the precision, 

%for three different environmental complexity. The vertical axis in the figure shows the speed of the robot and the horizontal axis shows the CPU voltage/frequency level. The colorbar shows the precision of the applications where the value 1.0 represents full precision state. The number inside each box shows the obtained power consumption in runtime. The boxes bolded via black shows the minimum obtained power for each combination of CPU volage/frequency and robot speed. 

%%%%%%%%%%%%%%%%%%%%%%%%%%%%%%%%%%%%

%The operation of performance management algorithm is explained better through an example. \NOTE{Here an example should be added}

%%%%%%%%%%%%%%%%%%%%%%%%%%%%%%%%%%%%
%%%%%%%%%%%%%%%%%%%%%%%%%%%%%%%%%%%%
%%%%%%%%%%%%%%%%%%%%%%%%%%%%%%%%%%%%
%%%%%%%%%%%%%%%%%%%%%%%%%%%%%%%%%%%%

\section{Experimental results}\label{sec:results}

\begin{figure*}[ht]
     \centering
    \begin{subfigure}[b]{0.32\textwidth}
         \centering
         \includegraphics[width=1\textwidth]{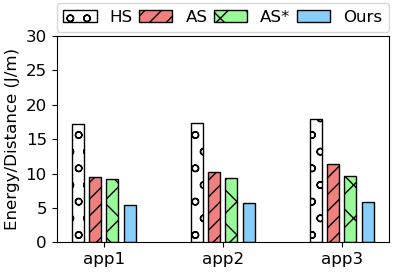}
         \caption{Low-complexity environment}
         \label{fig:apower_a}
     \end{subfigure}
     \hfill
     \begin{subfigure}[b]{0.32\textwidth}
         \centering
         \includegraphics[width=1\textwidth]{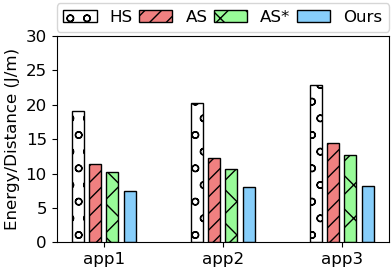}
         \caption{Medium-complexity environment}
         \label{fig:apower_b}
     \end{subfigure}
     \hfill
     \begin{subfigure}[b]{0.32\textwidth}
         \centering
         \includegraphics[width=1\textwidth]{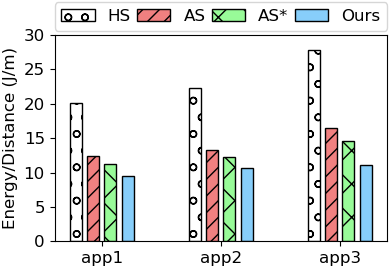}
         \caption{High-complexity environment}
         \label{fig:apower_c}
     \end{subfigure}
        \caption{Evaluation of the average energy per distance in three different environmental complexity.}
        \label{fig:apower}
            \vspace{-7pt}
\end{figure*}
% \begin{table}[t]
% \caption{Different obtained throughput for different control techniques. The average entropy, i.e., $Entropy$, and the number of generated events are the same for different techniques in each environment. }\centering
% \small
% \label{tab:exe}
% \begin{tabular}{|l|c|c|c|c|}
% \hline
% \multicolumn{2}{|l|}{}                 &\begin{tabular}[c]{@{}c@{}}Low\\ complexity\end{tabular}& \begin{tabular}[c]{@{}c@{}}Medium\\ complexity\end{tabular}& \begin{tabular}[c]{@{}c@{}}High\\ complexity\end{tabular} \\ \hline
% \multicolumn{2}{|l|}{avg. Entropy}          & 5.73         & 6.54            & 7.39          \\ \hline
% \multicolumn{2}{|l|}{Events {[}$10^6${]}} & 6.4          & 17.3            & 27.5          \\ \hline
% \multirow{4}{*}{\rotatebox[origin=c]{90}{\footnotesize{Throughput}}}   & \textbf{HS}     & 0.36           & 0.23               & 0.1             \\ \cline{2-5} 
%                               & \textbf{AS}     & 0.6           & 0.5              & 0.4            \\ \cline{2-5} 
%                               & \textbf{AS\textsuperscript{*}}    & 0.73           & 0.63             & 0.54            \\ \cline{2-5} 
%                               & \textbf{Ours}   & \textbf{1}           & \textbf{1}              & \textbf{1}            \\ \hline
% \end{tabular}
% \end{table}

The proposed controller has been implemented in C++ that is executed as a middle-ware in Linux OS user space at run-time. 
%\NOTE{here a paragraph stating that you developed a prototype of the controller in C running in user space on the top of Linux OS.}
%\NOTE{please comment a little bit better results by reporting some stats (x\% of improvement...) we have free space and the results section is quite short}
We run the proposed control algorithm for three different environments, i.e., 1) low, 2) medium-complexity, and 3) high-complexity.

\textbf{Effectiveness:} To demonstrate the effectiveness of the proposed approach we compared it against three different baseline schemes in the experiments: 1) while the controller uses the highest possible speed for the robot and highest possible voltage/frequency, i.e., $HS$, 2) while the controller uses a medium speed and highest possible voltage/frequency, i.e., $AS$, and 3) while the controller uses the medium speed with a medium voltage frequency, i.e., $AS^*$. Three applications are used as vision-based tasks, i.e., 1) image reconstruction that is shown by $app1$, 2) corner detection, $app2$, and 3) corner detection without filtering, $app3$, please see Section~\ref{sec:appl}. The overall number of generated events for each environment is the same for all of these control techniques, including the proposed method. The reason is that the environment is the same for all techniques\footnote{In macroscopic level, the amount of perceptual information for a mobile robot in a specific path is constant}. Figure~\ref{fig:thru} shows the average obtained throughput for different techniques. It can be seen that the proposed method obtains the highest average throughput in comparison with other methods. Here only the proposed controller can keep the throughput satisfiable for all the information environments. This is because of the fact that the motor speed affects the workload and should be tuned based on the workload of the system.

\begin{figure}[ht]
    \centering
     \begin{subfigure}[b]{0.4\textwidth}
         \centering
         \includegraphics[width=1\textwidth]{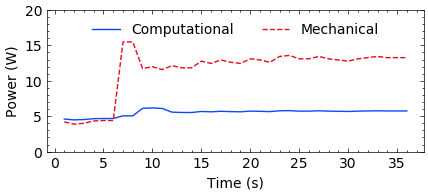}
         \caption{Low-complexity environment}
         \label{fig:et1}
     \end{subfigure}\\
     \begin{subfigure}[b]{0.4\textwidth}
         \centering
         \includegraphics[width=1\textwidth]{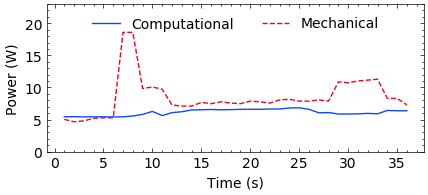}
         \caption{Medium-complexity environment}
         \label{fig:et2}
     \end{subfigure}\\
     \begin{subfigure}[b]{0.4\textwidth}
         \centering
         \includegraphics[width=1\textwidth]{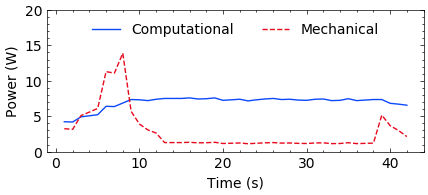}
         \caption{High-complexity environment}
         \label{fig:et3}
     \end{subfigure}
     \caption{Evaluation of the power consumption during the time for the three different scenarios.\label{fig:etX} }\centering
         \vspace{-7pt}
\end{figure}

\noindent\textbf{Efficiency.} To show the efficiency of our technique, we used the \textit{energy per meter} metric that shows the amount of energy consumption per distance unit for a mobile robot running a specific application. Figure~\ref{fig:apower} shows the overall energy consumption per meter (J/m) of the different techniques for the three different applications and in three different environments. As can be seen, the proposed approach saves the energy of traveling per distance unit significantly compared with the other methods. Our method is able to save an average of 50.5\%, 41\%, and 31\% of total energy in a low, medium, and high-complexity environment respectively.

%\NOTE{please check if the reference to the figure is ok.}
%\NOTE{do you think fig 7 is ok? you mix 2 different types of graphs}

\noindent\textbf{Stability.} To have a better understanding of the controller operation and to demonstrate the stability of the controller, the power and speed traces of two experiments in high, medium, and low-complexity environments, while $app1$ is executing, are shown in Figure~\ref{fig:etX}. As can be seen, at the beginning, the controller starts the hill-climbing process to find the appropriate speed and adjusts the CPU's voltage/frequency. This causes fluctuations in speed during the start-up time. However, for the rest of the travel, the controller adjusts the speed, and CPU power is quite steady. The reason is that environmental information usually does not change rapidly. This gives the hill-climbing algorithm an opportunity to adjust the CPU voltage/frequency efficiently.

%\noindent\textbf{Scalability.} As a final note, it is worth commenting the fact that, as shown by the experiments and in particular by Figure~\ref{fig:res}, the proposed hill-climbing strategy is effective due to the fact that the solution space is relatively small and is characterized by a concave surface of energy consumption. Therefore, an iterative search based on local moves shortly reaches the optimum. Thus, no scalability issue arises.

\noindent\textbf{Execution times.}
The average execution time of each module is as follows: the event-based camera module: 10 us; the frame-based camera module: 2.4 ms; and the Performance Management Unit module: 50 ms that is the average time for finding the optimal solution. As can be noticed, the execution time for the algorithm is acceptable for a macroscopic size robot with a normal change of speed rate. 

% \begin{table}[h]
% \centering
% \begin{threeparttable}
% \caption{ }\label{tab:exe}
% \begin{tabular}{l|c|c|c}
% \hline
%                      & low-textured & medium-textured & high-textured
%                     \\ \hline
% %%%%%%%%%%%%%%%%%%%%%%%%%%%%%%%%%%%%
% \textbf{avg. entropy} & 5.73 &6.54 &7.39\\
% \textbf{ no. events $[10^6]$}& 6.4 & 17.3&  27.5\\
% \textbf{ HS}  &16&  11& 9\\
% \textbf{ AS} & 23 & 21& 16\\
% \textbf{ AS\textsuperscript{*}} &20& 15& 12\\
% \textbf{ Ours} & 30 & 30 & 29\\
% \hline

% %%%%%%%%%%%%%%%%%%%%%%%%%%%%%%%%%%%%
% \end{tabular}
% \end{threeparttable}
% \end{table}

%We assume that the mobile robot runs event camera-based machine vision applications. Each application has predefined requirements that should be met at run-time. 

%\textit{test} the possible solutions for the (CPU voltage/frequency, robot speed) pair. 
%This setup provides a model for a \textit{cyber physical agent}~\cite{cyber-physical-agents}, where the robot, i.e., an agent, is continuously aware of its internal and external environment states to obtain a near-optimal energy-efficient solution.
%Experimental results with a robot mounting a Jetson TX2 board and a DAVIS-346 event-based camera show that the proposed control algorithm can save battery energy by an average of 50.5 \% in low complexity environment, 41\% in medium complexity environment, and 30\% in high complexity environment in comparison with naive controllers. 

\section{Related works}\label{sec:relatedworks}

The related works for energy-efficient robot control, or a swarm of robots, can be divided into two main categories: 1) those works that focus on energy-efficient computation for robotic applications, and 2) those works that focus on reducing the energy cost of the mechanical parts. From the computing perspective, several strategies have been used to control the instantaneous power and overall energy of the computation using different techniques acting at both architecture and application levels, such as DVFS~\cite{DVFS}, energy-efficient task mapping and scheduling~\cite{iccd-2014}, and software approximate computing~\cite{antonio}. The main idea in these works is to reduce the computing power/energy as much as possible while keeping the quality of service (in terms of the provided throughput) satisfactory. On the other hand, and from the robotics perspective, there have been several researches to optimize the energy cost and battery utilization of the mechanical parts that are ranging from conceptual studies such as bio-inspired theories based on least action principle (LAP)~\cite{fox-entropy}, to optimizing mechanical motion algorithms, e.g., reducing locomotion cost of different parts of the robot~\cite{c3, LM2} and energy-efficient path planning~\cite{b2, AAA}.

To the best of our knowledge, all of the studies in these two areas are considering the optimization of the computing and mechanical parts independent from each other, i.e., co-optimization has not been investigated in practice. In~\cite{fox-entropy}, the authors have proposed a \textit{new theory} in multi-robot environment in which computing and mechanical parts are considered together under the umbrella of overall \textit{entropy measurement} of the system. It is stated that the main goal of an autonomous system should be to save energy by reducing the entropy as much as possible. In~\cite{cyber-physical-agents}, the author has proposed an \textit{architectural characterization} for cyber-physical agents that consists of \textit{internal and external environments}, and by evaluating measurements from these two environments the controller should decide the values of \textit{internal and external actuators}. In this paper, the internal and external environments correspond to the computing platform and the robot environment respectively, and, internal and external actuators correspond to voltage/frequency scaling for the computing unit and controlling the robot speed respectively.

\section{Conclusions}\label{sec:conclusions}
In this paper, a novel control approach is proposed to intelligently tune the CPU voltage/frequency and motor voltage of a mobile robot running vision-based applications. The proposed approach uses the complexity of the environment, instantaneous power of the motor, and CPU to compute the overall energy consumption of the system and manipulate the CPU voltage/frequency and the motor voltage of the robot at run-time. 
%We have shown that in mobile robot platforms running heavy vision-based applications, the contribution of the computing energy on the overall energy consumption of the robot is significant. Moreover, we demonstrated a measurement technique through which the effect of the decision-making in the mechanical part on the computing workload can be estimated. This is done via the relationship between the complexity of the environment, i.e, the entropy computed from the captured images and the speed of the robot. 
A run-time hill-climbing algorithm has been proposed to find the near-optimal energy-efficient solution for the controller. Experimental results show that by utilizing our method a mobile robot equipped with a Jetson TX2 board can save an average of 50.5\% of the energy in a low-complexity environment, 41\% in a medium-complexity environment, and 30\% in a high-complexity environment. As future work, we intend to extend our approach by considering the energy consumption of the path planning module.

\section*{Acknowledgment}
This work has been financially supported by the Academy of Finland funded projects 335512 - ADAFI (Adaptive-Fidelity Digital Twins for Robust and Intelligent Control Systems) and 330493 - AURORA (Autonomous Performance Management in Digital Manufacturing), and  by Nokia Jorma Ollila Grant.

\bibliographystyle{IEEEtran}
\bibliography{ref}
\end{document}